\documentclass{article}

\PassOptionsToPackage{numbers, compress}{natbib}
    \usepackage[final]{tackling_climate_workshop_style}

\usepackage[utf8]{inputenc} % allow utf-8 input
\usepackage[T1]{fontenc}    % use 8-bit T1 fonts
\usepackage{hyperref}       % hyperlinks
\usepackage{url}            % simple URL typesetting
\usepackage{booktabs}       % professional-quality tables
\usepackage{amsfonts}       % blackboard math symbols
\usepackage{nicefrac}       % compact symbols for 1/2, etc.
\usepackage{microtype}      % microtypography
\usepackage{graphicx}
\usepackage{float}
\usepackage{subfig}

\title{Utilizing a Geospatial Foundation Model for Coastline
Delineation in Small Sandy Islands}

\author{%
  Tishya Chhabra \\
  \texttt{tishyac3.141@gmail.com} \\
  \And
  Manisha Bajpai \\
  \texttt{mbajpai@gmail.com} \\
  \AND
  Walter Zesk \\
  \texttt{wzesk@mit.edu} \\
  \And
  Skylar Tibbits \\
  \texttt{stibbits@mit.edu} \\
}

\begin{document}

\maketitle

\begin{abstract}
  We present an initial evaluation of NASA and IBM’s Prithvi-EO-2.0 geospatial foundation model for the task of shoreline delineation on small sandy islands using satellite imagery. We curated and labeled a dataset of 225 multispectral Sentinel-2 images of two Maldivian islands, which we publicly release, and fine-tuned both the 300M and 600M parameter versions of Prithvi-EO-2.0 on training subsets ranging from 5 to 181 images. Our experiments show that even with as few as 5 training images, the models achieve high segmentation performance (F1 > 0.94, IoU > 0.79). We observed minimal differences between the 300M and 600M models, suggesting that the larger model’s additional computational cost may not be justified for this task. The results demonstrate the strong transfer learning capability of Prithvi, underscoring the potential of such models to support coastal monitoring in data-poor regions.
\end{abstract}

\section{Introduction}
For over seven years, the Self-Assembly Lab at MIT has pioneered the Growing Islands project to address sea-level rise and coastal erosion by strategically placing geometric structures underwater \cite{lab_growing_nodate}. Piloted in the Maldives, these structures have facilitated the natural accumulation of sand, forming up to meter-long sandbars. A key question vital to the project is deciding which parts of the shoreline to address first. Analyzing decades of satellite images would allow the team to thoroughly understand short-term and long-term shoreline dynamics. This type of coastal analysis is vital not only to this project in the Maldives, but to coastlines around the world, so we can sustainably protect and rebuild coastal communities and habitats. 

Delineating coastlines from satellite images is the first step to understanding shoreline dynamics. However this remains a challenge, especially with small islands. Several deep learning methods have been applied \cite{mcallister_multispectral_2022, vos_benchmarking_2023, ramesh_monitoring_2020, toure_shoreline_2019, lv_research_2024, dang_application_2022, seale_coastline_2022, adusumilli_predicting_2024}, but most are trained for specific coastal regions, leaving small sandy islands underrepresented. For a number of Maldivian islands, tools such as CoastSat and CoastSeg \cite{vos_coastsat_2019, sharon_fitzpatrick_coastseg_2024} struggle to differentiate between the shoreline versus the atoll that the island sits on (see Appendix for details). Modern computer vision models such as SAM and YOLO cannot process multispectral imagery, forcing a reduction in the rich geospatial information embedded in satellite images. Finally, the most accessible satellite images, both in terms of cost and time range, are low resolution (such as 10-meter or 30-meter), posing an additional challenge. Obtaining higher resolution imagery is not only costly, but often covers a short amount of time, prohibiting historical analysis.

Geospatial foundation models (GFM) have emerged over the past few years. Pretrained on vast amounts of earth observation data, they enable transfer learning to a variety of geospatial tasks, even with sparse labeled data \cite{vatsavai_geospatial_2024, xiong_neural_2024}. One of the most notable geospatial models is Prithvi-EO-2.0, released by NASA and IBM in December 2024 \cite{szwarcman_prithvi-eo-20_2025}. Initial applications and benchmarking of the model, such as \cite{lambhate_finetuning_2024, muszynski_fine-tuning_2024, li_assessment_2023, hsu_geospatial_2024, blumenstiel_multi-spectral_2024}, offer a glimpse into Prithvi-EO-2.0’s capabilities, but there still remains opportunity for thorough evaluation. Furthermore, to our knowledge, no GFM has yet been applied on the vital task of shoreline delineation.

In this paper, we aim to bridge these gaps by testing Prithvi-EO-2.0 300M and 600M on the task of delineating shorelines from lower resolution satellite images using image segmentation. Furthermore, we evaluate the real-world applicability of Prithvi by examining the relationship between training dataset size and performance. We also release a curated dataset of 225 Sentinel-2 images taken at 10 meter resolution of two islands in the Maldives, Fuvamulah and Madhirivaadhoo, paired with hand-labeled ground truths, which can be found at this link: \url{https://github.com/tishyac3141/Maldives-Image-Segmentation-Sentinel-2-Dataset}. 
This work is posed as an initial assessment of the applicability of geospatial foundation models like Prithvi  to coastline delineation. 

\section{Prithvi Overview}
Prithvi-EO-2.0 was trained on an extensive global dataset of 4.2 million satellite images from NASA’s Harmonized Landsat Sentinel-2 (HLS) archive spanning a decade \cite{szwarcman_prithvi-eo-20_2025}. These satellite images contain six optical bands: Red, Green, Blue, NIR, SWIR1, and SWIR2. The pretraining data set was built using an extensive sampling strategy to ensure geographical diversity during pretraining. Prithvi’s key architectural component is the Vision Transformer (ViT)\cite{dosovitskiy_image_2021}, utilizing the Masked Autoencoder approach as its pre-training strategy \cite{he_masked_2021}. Prithvi-EO-2.0 was pre-trained for 400 epochs, with the 300M parameter model utilizing 80 GPUs and the 600M model utilizing 240 GPUs \cite{szwarcman_prithvi-eo-20_2025}. Because of these features, Prithvi-EO-2.0 is well equipped to easily and quickly capture patterns and map Earth's features to a rich embedding space.

\section{Methodology}

\subsection{Data}
We collected 225 multispectral images of two Maldivian islands at 10m resolution from Sentinel-2. Example images can be reviewed in section A.2 of the Appendix. The GeoTIFFs were retrieved using Google Earth Engine via CoastSat\cite{vos_coastsat_2019}, which provides a standardized preprocessing pipeline optimized for coastal analysis. Each of the 225 images were then labeled using Kili Technology where we manually traced the shoreline in each TIFF image. Pixels inside the boundary were labeled as land and pixels outside as water. 

The final 225-image dataset was split into a training set of 181 images, and a validation and test set of 22 images each. Additionally, to examine the relationship between training dataset size and performance, we created smaller subsets of the training dataset by choosing images randomly from the full training set. The subsequent training dataset sizes are as follows: 5 images, 10 images, 25 images, 50 images, 75 images, 100 images, 125 images and 150 images. The validation and test datasets remained the same 22 images, to ensure that only dataset size was being tested. 

\subsection{Experimental Setup}
NASA and IBM released two versions of Prithvi:  300M parameters and 600M parameters. We tested both versions by instantiating separate versions of the 300M and 600M model for each training dataset size (ie 5 images, 10 images, 25 images, etc) and then fine tuning for 30 epochs. They were validated and tested on the same validation and test sets of 22 images each. 

We froze the ViT backbone during fine tuning, an intentional decision made to simulate real-world constraints on training time and computational resources. Furthermore, since our data consists of Sentinel-2 images, on which the encoder has already been extensively pre-trained on, we expect transfer learning to be largely effective.

Further details on our setup can be found in section A.2.1 of the Appendix. 

\section{Results}
\begin{table}[H]
  \caption{Performance of Prithvi 300M and 600M models at varying training dataset sizes.}
  \label{tab:prithvi-results}
  \centering
  \begin{tabular}{lcccc}
    \toprule
    Training Dataset Size & \multicolumn{2}{c}{IoU} & \multicolumn{2}{c}{F1} \\
    \cmidrule(r){2-3} \cmidrule(r){4-5}
    & Prithvi 300M & Prithvi 600M & Prithvi 300M & Prithvi 600M \\
    \midrule
    5 images   & 0.8509 & 0.7977 & 0.9612 & 0.9429 \\
    10 images  & 0.8720 & 0.8927 & 0.9664 & 0.9743 \\
    25 images  & 0.9423 & 0.9311 & 0.9870 & 0.9845 \\
    50 images  & 0.9427 & 0.9475 & 0.9871 & 0.9880 \\
    75 images  & 0.9470 & 0.9538 & 0.9881 & 0.9896 \\
    100 images & 0.9536 & 0.9577 & 0.9896 & 0.9905 \\
    125 images & \textbf{0.9593} & \textbf{0.9616} & \textbf{0.9908} & \textbf{0.9913} \\
    150 images & 0.9570 & 0.9529 & 0.9904 & 0.9894 \\
    181 images & 0.9529 & 0.9553 & 0.9894 & 0.9899 \\
    \bottomrule
  \end{tabular}
\end{table}

Our results show that Prithvi-EO-2.0 applies well to the task of delineating coastlines from satellite images, with all our scores summarized in Table 1. Section A.2 in the Appendix contains plots further visualizing our results as well as sample outputs from the fine tuned models. There are some additional aspects to note about our results:

1) Our highest F1 score is Prithvi 600M at 0.9912, with Prithvi 300M closely following at 0.9908, both fine-tuned on 125 images. Our lowest F1 score is Prithvi 600M at 0.9429, fine-tuned at 5 images. 

2) When the training dataset size is very small at 5 images, Prithvi-EO-2.0 300M does significantly better (0.8509 IoU) than Prithvi-EO-2.0 600M (0.7977 IoU) despite being the smaller model.

3) Generally, the difference in performance between Prithvi-EO-2.0 300M and Prithvi-EO-2.0 600M is marginal, making it questionable whether the extra computational load is worth using the 600M parameter model.

4) An extremely small training dataset yields decent performance, underscoring that using Prithvi even with small labeled datasets could be useful and worth the effort. 

5) Given the size of our dataset, performance quickly levels out around 10-15 epochs (see figure in Appendix). Prithvi-EO-2.0 600M version seems to stabilize more quickly.

Although the models may exhibit some degree of overfitting, given that the training dataset was limited to two islands in the Maldives, this is not necessarily detrimental in the context of foundation model adaptation. Since foundation models are designed to be fine-tuned for specific downstream tasks, a certain level of overfitting is acceptable provided the model generalizes well within the intended scope — in this case, across Maldivian islands. Prithvi can subsequently be fine tuned for each coastal region.

\section{Conclusion and Future Work}
Island segmentation with Prithvi-EO-2.0 was remarkably accurate, even when fine-tuned on only a few labeled images. Our study demonstrates the untapped potential of Prithvi for shoreline segmentation. This is a critical development for two reasons. One, it validates the new paradigm shift of fine-tuning with small datasets, alleviating the more traditionally high costs and making geospatial AI a more accessible tool in a coastal setting. Two, our validation shows lower resolution imagery, like our dataset taken at 10 meters, can in fact be used to reliably extract shorelines. This is vital to constructing a complete historical record reflective of true change over time. Current toolkits struggle with small islands, as seen in section A.1 of the Appendix; however Prithvi worked very well in this context, unlocking a new approach for otherwise neglected regions. With this study serving as a starting point, there are a number of future directions to explore, which we detail in the Appendix. But the impact of better coastline monitoring in a small sandy island setting are immediate. Being able to delineate coastlines from images at scale would allow for informed decision-making so coastal communities can remain resilient in the face of change.

\bibliographystyle{unsrt}
\bibliography{references}

\clearpage
\appendix
\section{Appendix}
\addcontentsline{toc}{section}{Appendix}
\setcounter{figure}{0}
\renewcommand{\thefigure}{A\arabic{figure}}

\subsection{Previous Work}
\begin{figure}[H]
  \centering
  \includegraphics[width=0.55\linewidth]{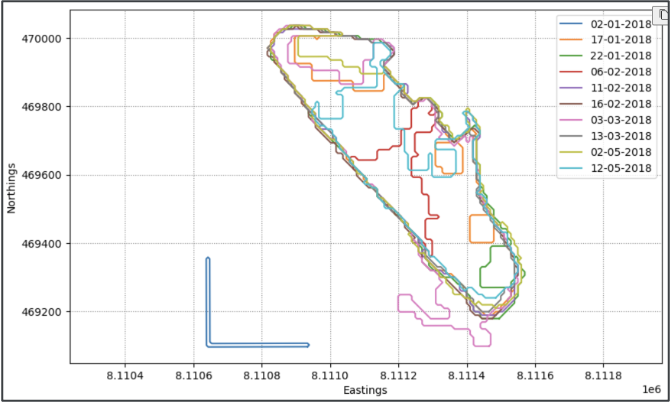}
  \caption{This is a run where we used CoastSat, a well-known coastline delineation toolkit, end to end to extract coastlines from an island in the Maldives. It can clearly be seen that the boundaries are varying extremely with tons of artifacts, even though the island shoreline itself is not changing so dramatically. Thus, these shorelines are not usable, highlighting the gap that our implementation fills when it comes to shoreline extraction.}
\end{figure}

\subsection{Data and Experimental Setup}
\begin{figure}[H]
  \centering
  \subfloat[Madhirivaadhoo]{\includegraphics[width=0.4\linewidth]{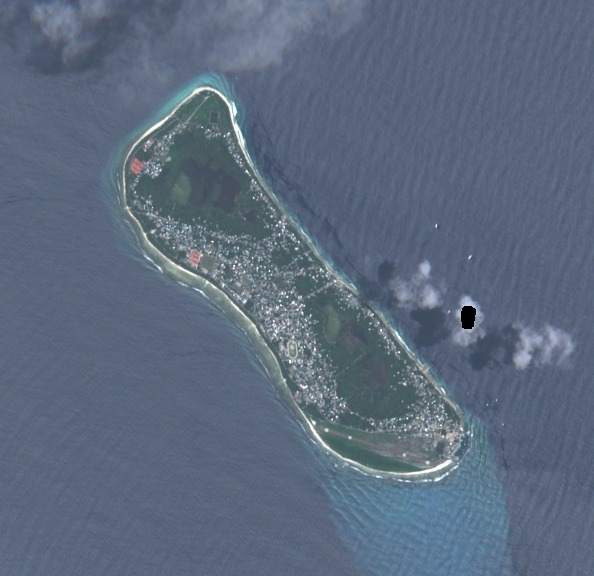}\label{fig:f1}}
  \hfill
  \subfloat[Fuvahmulah]{\includegraphics[width=0.4\linewidth]{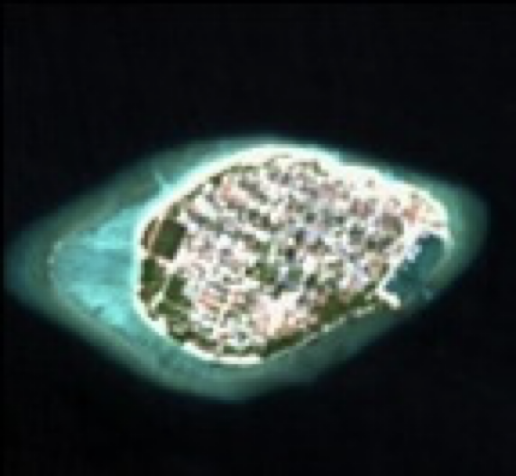}\label{fig:f2}}
  \caption{These are sample images of the two islands that our dataset is comprised of. The images were GeoTIFFS taken from Sentinel-2 and retrieved using Google Earth Engine. The images contain 6 optical bands -- R, G, B, Near Infrared (NIR), and Shortwave Infrared (SWIR) 1 and 2.}
\end{figure}

\subsubsection{Experimental Setup}

The images were resized to 224x224 pixels, and our batch size was set to 4 images. The decoder employed a U-Net–style architecture with skip connections and channel widths of 512, 256, 128, and 64, followed by a segmentation head with a dropout rate of 0.1. The model was trained to predict two classes, water and land, using cross-entropy loss. Fine tuning was performed on a single GPU using bfloat16 mixed-precision arithmetic to reduce memory usage and accelerate computation. We optimized the model using the AdamW optimizer with a learning rate of $10^{-4}$ for a maximum of 30 epochs. The best model checkpoint was selected based on the highest validation Jaccard Index (IoU). Training progress was logged using TensorBoard, and checkpoints were saved after each epoch. 

\subsection{Results}
\begin{figure}[H]
  \centering
  \subfloat[F1 Scores vs Training Dataset Size.]{\includegraphics[width=0.5\linewidth]{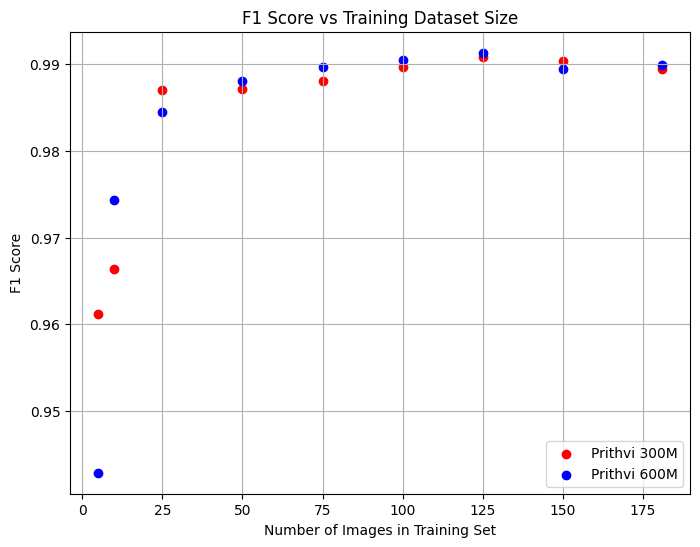}\label{fig:f1}}
  \hfill
  \subfloat[IoU Scores vs Training Dataset Size.]{\includegraphics[width=0.5\linewidth]{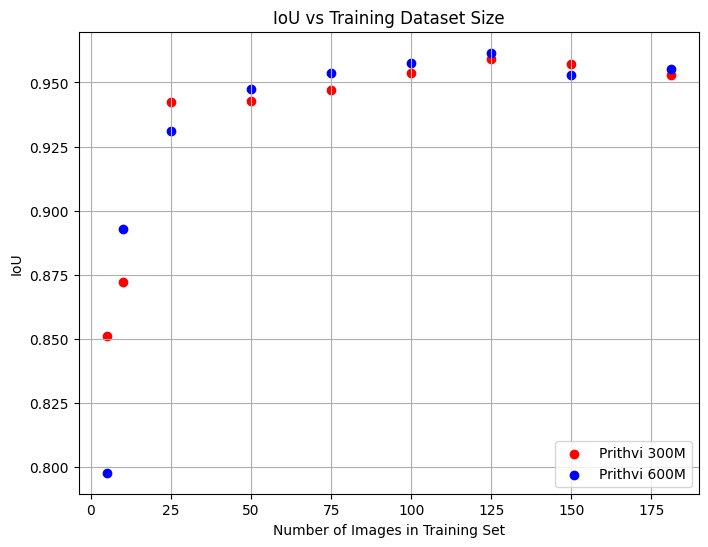}\label{fig:f2}}
  \caption{Plotted Results of Prithvi 300M (Red) and Prithvi 600M (Blue), showing the relationships between performance and training dataset size. The x-axis has each of our sub training dataset sizes, while the y-axis has the range of score values, with the left plot showing F1 scores and the right plot showing IoU scores.}
\end{figure}

\begin{figure}[H]
  \centering
  \includegraphics[width=0.98\linewidth]{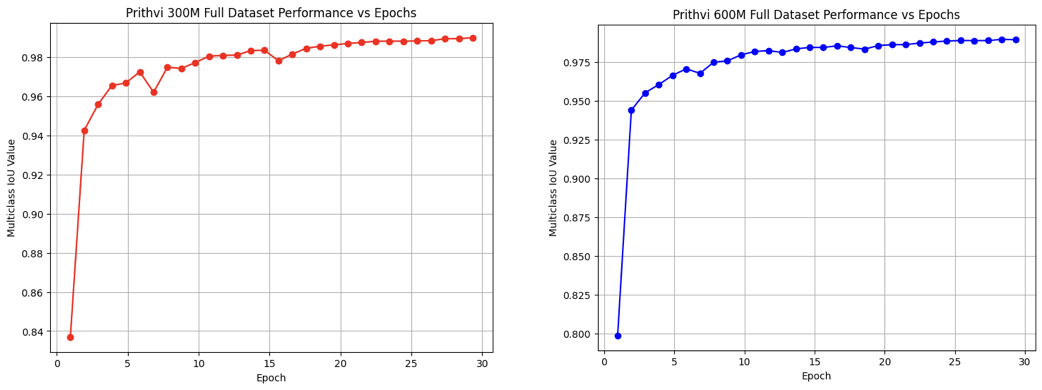}
  \caption{These graphs show the performance of Prithvi 300M and Prithvi 600M, specifically the IoU scores, over the epochs. We can see performance levels out fairly quickly, around 10-20 epochs, with Prithvi 600M stabilizing more quickly.}
\end{figure}

\begin{figure}[H]
  \centering
  \includegraphics[width=0.85\linewidth]{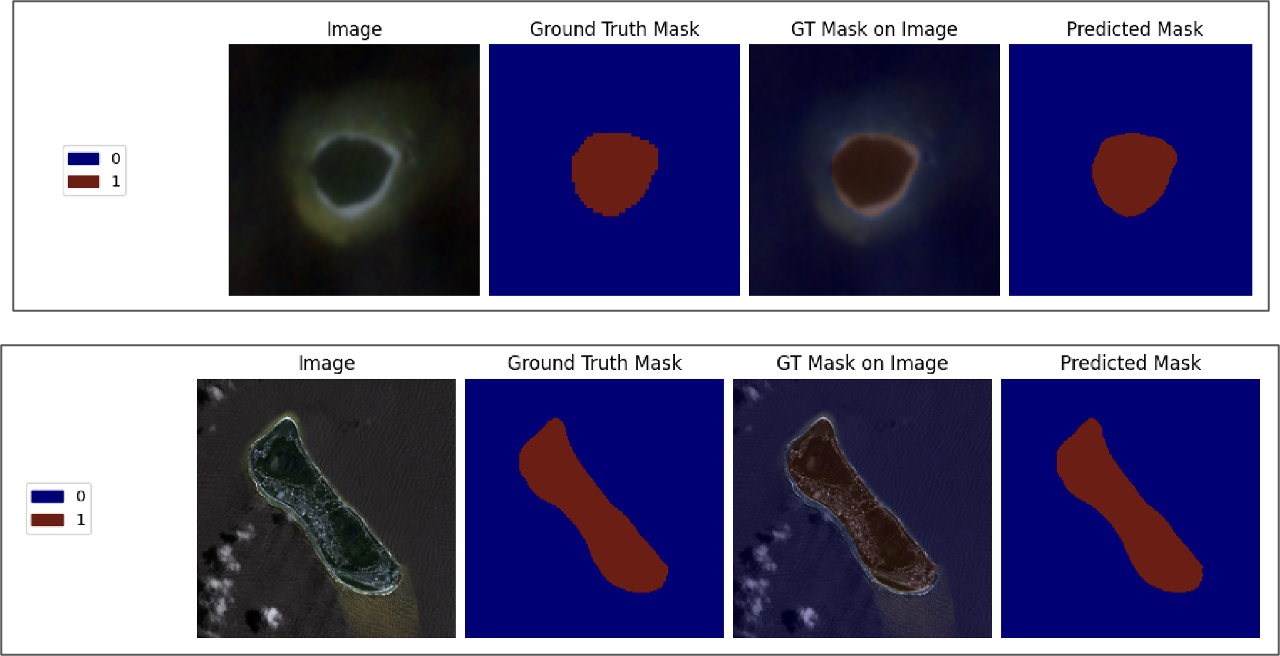}
    \caption{Figure A4 and Figure A5 are sample outputs of the fine-tuned versions of Prithvi. This Figure A4 shows a sample output of Prithvi 300M. Below, Figure A5 shows a sample output of the 600M version. Subtle differences can be seen between both of their predicted masks.}
\end{figure}

\begin{figure}[H]
  \centering
  \includegraphics[width=0.85\linewidth]{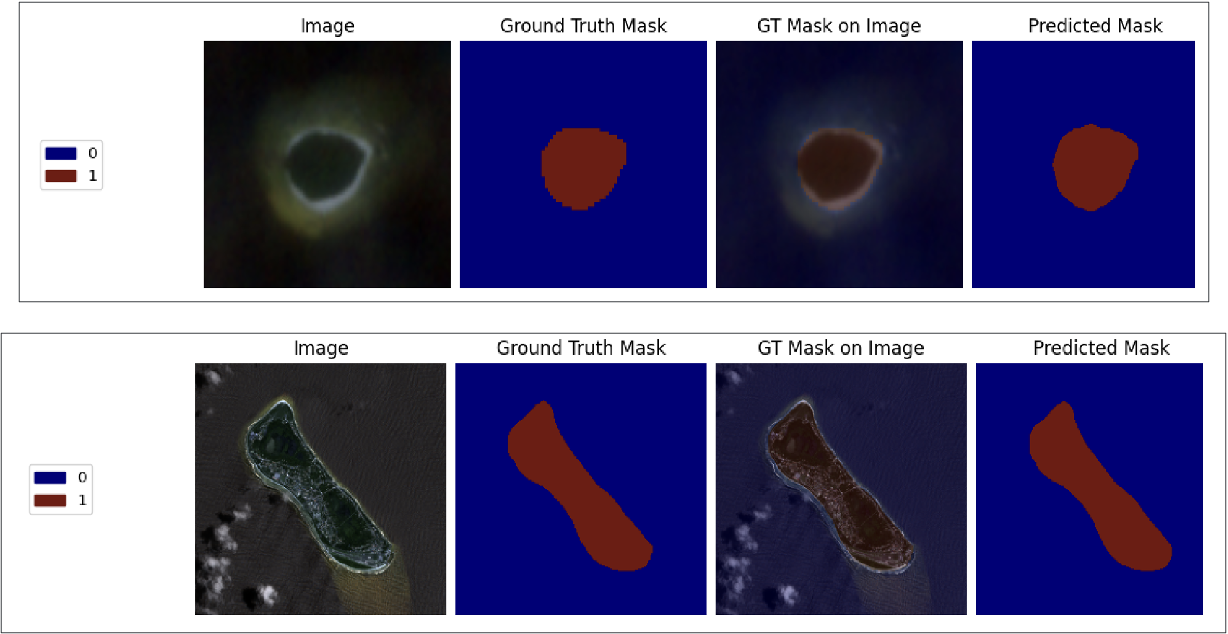}
  \caption{Figure A4 and Figure A5 are sample outputs of the fine-tuned versions of Prithvi. Figure A4 above shows a sample output of Prithvi 300M, while Figure A5 shows a sample output of Prithvi 600M. Subtle differences can be seen between both of their predicted masks.}
\end{figure}

\subsection{Future Work}
With this study serving as a starting point, there are a number of future directions to explore. From our results alone, there are a number of addressable questions, such as why performance peaks at 125 images, what features in the images cause performance differences, and how this approach compares with other segmentation approaches and models. Expanding our analysis beyond two islands in the Maldives would also provide deeper and more generalizable analysis and results. In terms of further evaluating Prithvi-EO-2.0's applicability to this task, along with the 300M and 600M parameter versions, Prithvi-EO-2.0 also has "TL" versions of those models where satellite metadata can be included, which would be worth testing. Finally, incorporating other data modalities, whether into Prithvi-EO-2.0 or coupled with other modality-specific models, would be an interesting exploration into more accurately delineating coastlines. Given the changing tides, changing beaches, and changing sea levels, image segmentation is an important first step, but cannot be the sole method for extracting accurate shorelines.

\end{document}